\documentclass{llncs} %

\usepackage{float}
\usepackage{graphicx}
\usepackage{wrapfig}

\usepackage[labelfont=bf]{caption}
\usepackage{color}
\usepackage{amsmath} 
\usepackage{mathtools} 
\usepackage{amsfonts} 
\usepackage{amssymb}
\usepackage[normalem]{ulem}

\usepackage{array} 
\usepackage{booktabs}
\usepackage{rotating}
\usepackage{multirow}
\usepackage{multicol} 
\usepackage{lscape}
\usepackage{subfig}
\usepackage{comment}

\usepackage[export]{adjustbox}

\usepackage[ruled,vlined]{algorithm2e}

\DeclareMathOperator*{\argmin}{arg\,min}  

\usepackage[export]{adjustbox}

\usepackage{xcolor,soul}
\definecolor{pred}{HTML}{ffe699}
\definecolor{gt}{HTML}{d5c3d2}

\usepackage[colorlinks=true,linkcolor=blue,citecolor=blue]{hyperref}

\usepackage{tikz,xcolor,hyperref}

\definecolor{lime}{HTML}{A6CE39}
\DeclareRobustCommand{\orcidicon}{
	\begin{tikzpicture}
	\draw[lime, fill=lime] (0,0) 
	circle [radius=0.16] 
	node[white] {{\fontfamily{qag}\selectfont \tiny ID}};
	\draw[white, fill=white] (-0.0625,0.095) 
	circle [radius=0.007];
	\end{tikzpicture}
	\hspace{-2mm}
}

\foreach \x in {A, ..., Z}{\expandafter\xdef\csname orcid\x\endcsname{\noexpand\href{https://orcid.org/\csname orcidauthor\x\endcsname}
			{\noexpand\orcidicon}}
}
			
\definecolor{darkgreen}{rgb}{0.53, 0.66, 0.42}

\begin{document}

\title{One Representative-Shot Learning Using a Population-Driven Template with Application to Brain Connectivity Classification and Evolution Prediction}

\titlerunning{One Representative-Shot Learning Using a Population-Driven Template}  

\author{Umut Guvercin\index{Guvercin, Umut} \and  Mohammed Amine Gharsallaoui\index{Gharsallaoui, Mohammed Amine} \and Islem Rekik\orcidA{} \index{Rekik, Islem}\thanks{corresponding author: irekik@itu.edu.tr, \url{http://basira-lab.com}}}

\institute{BASIRA Lab, Faculty of Computer and Informatics, Istanbul Technical University, Istanbul, Turkey}

\authorrunning{Umut Guvercin}

\maketitle              

\begin{abstract}
Few-shot learning presents a challenging paradigm for training discriminative models on a few training samples representing the target classes to discriminate. However, classification methods based on deep learning are ill-suited for such learning as they need large amounts of training data --let alone \emph{one-shot} learning. Recently, graph neural networks (GNNs) have been introduced to the field of network neuroscience, where the brain connectivity is encoded in a graph. However, with scarce neuroimaging datasets particularly for rare diseases and low-resource clinical facilities, such data-devouring architectures might fail in learning the target task. In this paper, we take a very different approach in training GNNs, where we aim to \emph{learn with one sample and achieve the best performance} --a formidable challenge to tackle. Specifically, we present the first one-shot paradigm where a GNN is trained on a \emph{single population-driven template} --namely a connectional brain template (CBT). A CBT is a compact representation of a population of brain graphs capturing the unique connectivity patterns shared across individuals. It is analogous to brain image atlases for neuroimaging datasets. Using a \emph{one-representative} CBT as a training sample, we alleviate the training load of GNN models while boosting their performance across a variety of classification and regression tasks. We demonstrate that our method significantly outperformed benchmark one-shot learning methods with downstream classification and time-dependent brain graph data forecasting tasks while competing with the ``train on all'' conventional training strategy. Our source code can be found at \url{https://github.com/basiralab/one-representative-shot-learning}.


\end{abstract}
 
\section{Introduction}

Deep learning models have achieved remarkable results across different medical imaging learning tasks such as segmentation, registration and classification \cite{shen2017deep,yi2019generative}. Despite their ability to extract meaningful and powerful representations from labelled data, they might fail to operate in a frugal setting where the number of the samples to train on is very limited. Besides, training such data-devouring architectures is computationally expensive and might not work on scarce neuroimaging datasets particularly for rare diseases \cite{schaefer2020use} and in countries with low-resource clinical facilities \cite{piette2012impacts} --limiting their generalizibility and diagnostic impact. Thus, deep networks seem less powerful when the goal is to learn a mapping classification or regression function on the fly, from only a few samples. Such problem is usually remedied by the few-shot learning (FSL) paradigm \cite{kadam2018review,sun2019meta,li2020concise}, where only few labeled samples are used to learn a new concept that can generalize to unseen distributions of testing samples. While training with lesser amount of labelled data reduces the cost in terms of computational power and required time, it also overcomes the data scarcity issue. Several works developed novel ways of leveraging FSL in medical image-based learning tasks.  For instance, \cite{zhao2019data} presented a learning-based method that is trained with few examples while leveraging data augmentation and unlabeled image data to enhance model generalizability. \cite{mondal2018few} also leveraged unlabeled data for segmenting 3D multi modal medical images. Their implementation of semi-supervised approach with generative adversarial networks obviated the need of pre-trained model. \cite{li2020difficulty} used the meta-train data from common diseases for rare disease diagnosis and tackled the low-data regime problem while leveraging meta-learning. However, such works resort to data augmentation strategies or generative models to better estimate the unseen distributions of the classes to discriminate. However, generating \emph{real} and \emph{biologically sound} data samples, particularly in the context where the ultimate goal is disease diagnosis and biomarker discovery, becomes a far cry from learning on a few samples \emph{alone}.  

In this paper, we take a very different approach and ask whether we can \textbf{learn with one sample and achieve the best performance without any augmentation} --a formidable challenge to tackle. Our approach aims to train FSL models on a single sample that is \emph{well-representative} of a particular class to learn--demonstrating that \emph{one is enough} if it captures well the unique traits of the given class. Our one-representative shot model does not require costly optimization and is efficient in one go. Deep learning methods have proven their efficiency in many applications. However, these methods are less successful when dealing with non-Euclidian data such as graphs and manifolds \cite{bronstein2017geometric}. To circumvent this limitation, many studies have proposed graph neural networks (GNNs), which extend the efficiency of deep learning to a broader range of data structures such as graphs \cite{kipf2016semi}. Particularly, GNNs have been widely used to tackle many problems in network neuroscience as detailed in a recent review paper \cite{bessadok2021graph}. So far, we also note that there is a limited number of works where FSL was adopted to train GNNs. For example, \cite{garcia2017few} designed a GNN \cite{gori2005new} architecture trained on a few shots. It aims to learn a complete graph network where nodes store features and class information. \cite{cheng2020attentive} proposes attentive GNN model for few-shot learning that can mitigate the over-fitting and over-smoothing issues in deep GNN models. However, to the best of our knowledge there is no study that focuses on training GNN models with \emph{a single shot} and where each sample is encoded as a whole graph. Recent adversarial GNN models have been tailored for brain graph evolution prediction from a single timepoint such as EvoGraphNet \cite{nebli2020deep} or brain graph classification \cite{bi2020gnea}. However, such architectures cross-pollinating the field of network neuroscience where the brain is represented as a graph encoding the pairwise interactions between anatomical regions of interest (ROIs), are still data-hungry and needy for high computational facilities.

\begin{figure}[ht!]
\includegraphics[width=12cm]{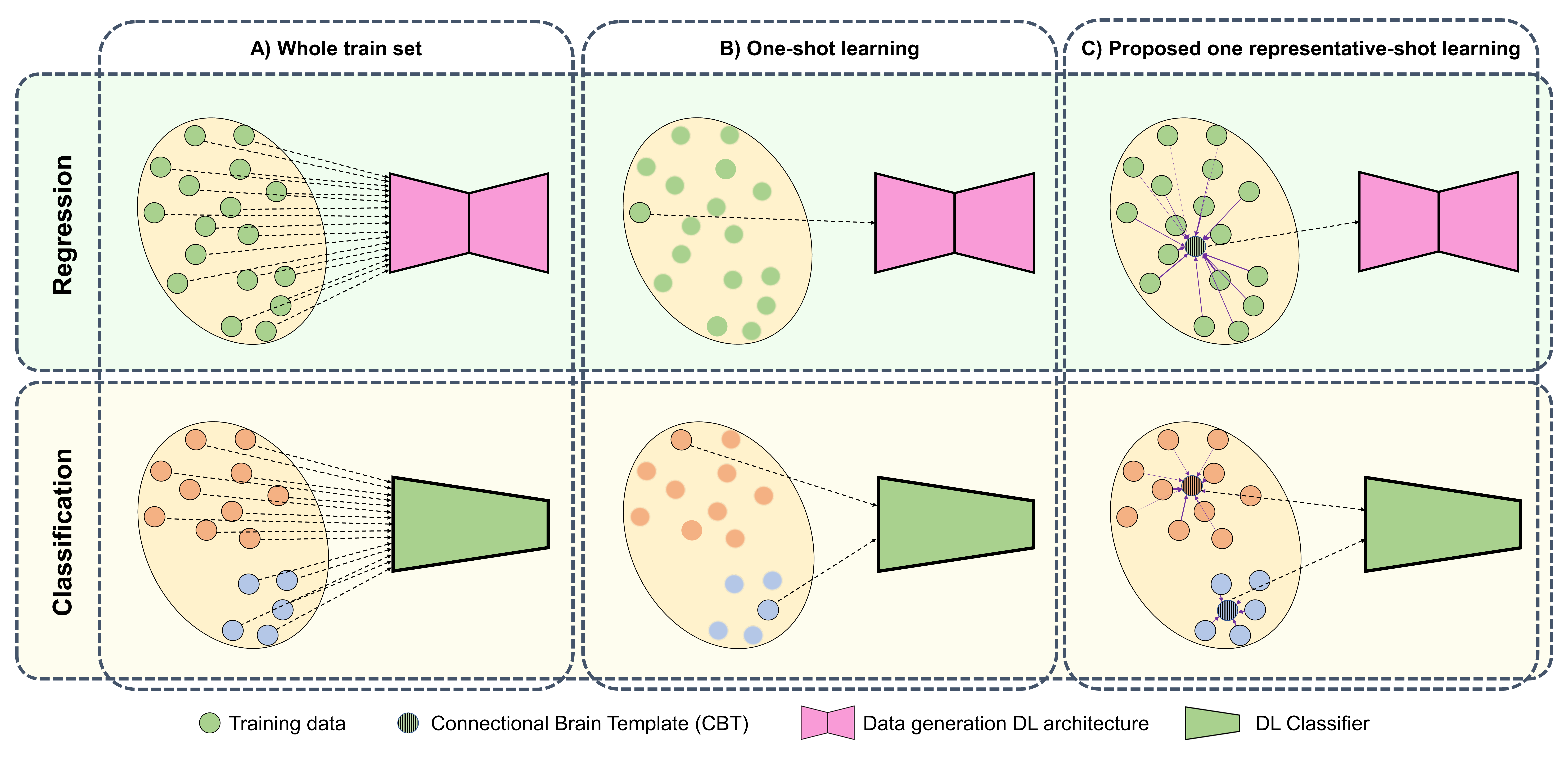}
\caption{\emph{Proposed one representative-shot learning compared to the conventional `train on all' strategy and one-shot learning.} \textbf{A) Whole train set.} The learned (regressor or classifier) is trained on the whole set. \textbf{B) One-shot learning.} The learner is trained with only on one sample. \textbf{C) Proposed one representative-shot learning.} The regressor is trained with the representative sample (i.e., CBT) of the whole population. The classifier is trained with the CBTs of each class.}
\label{mainfig}
\end{figure}

In order to train a GNN with a single sample, one faces the diversity of selecting samples and creating new one(s) or possibly both. Since the lack of data increases the importance of the quality and diversity of the training set, randomly selecting samples is neither viable nor stable. Clearly, learning from a single exemplar is only possible given sufficient prior knowledge on the learning domain. The main challenge is then to find a single sample that captures the heterogeneous distribution of the class to learn. Here we propose to leverage the emerging concept of a connectional brain template \cite{rekik2017estimation,dhifallah2020,gurbuz2020deep},  where a population of brain graphs is encoded in a compact connectivity matrix capturing the unique connectivity patterns shared across individuals of a given population. This can be regarded as network `atlas' by analogy to brain image template such as the Montreal Neurological Institute (MNI) template \cite{brett2001}. We propose to use such a compact and universal sample (i.e., CBT) to train GNN models for brain graph classification and time-dependent generation tasks. In opposition to the state-of-the-art methods which are data-hungry and needy for high computational facilities, our proposed approach can learn from one sample using the \textbf{paradigm-shift} we are creating through the \textbf{one-representation training} of deep learning models. The principle introduced is highly innovative and fully agnostic to the deployed methods.  Inspired from the CBT concept, this pioneering work can drive the whole field of geometric deep learning in a completely new direction related to the optimization of the amount of data needed to train a GNN model for a specific task. In addition to the computational gain, we  achieved striking results compared to other benchmark one-shot learning methods and a competitive performance to the ``train on all'' conventional training strategy across different connectomic datasets and learning tasks.

\section{Proposed Method}

\textbf{Overview.} In this section, we detail the foundational principles of the proposed representative one-shot learning method (\textbf{Fig.}~\ref{mainfig}) in comparison to training strategies. The first benchmark is resourceful (\textbf{Fig.}~\ref{mainfig}-A) as it consumes all available training data. In this case, it is generally assumed that the available training samples are well distributed across possible data domain. The second benchmark is the conventional one-shot learning where we only train our method on one sample of the data (\textbf{Fig.}~\ref{mainfig}-B). When we dive deeper into concrete implementations of such strategies, we face a peculiar challenge: the deep model performance can be highly sensitive to training and test set distribution shifts via local or global perturbations. This can be remedied by training on a single sample encapsulating sufficient prior knowledge on the learning domain, thereby allowing generalizability to unseen samples. Our proposed one representative-shot learning method (\textbf{Fig.}~\ref{mainfig}-C) supports such hypothesis where we investigate the potential of one shot (i.e., a CBT) in training well different learning models. Here, we focus on two different target learning tasks to evaluate our method. For regression, we use a CBT sample to represent the whole population; for classification, we use one class-specific CBT sample to represent each class.

\textbf{The CBT as one representative shot.} A well-representative connectional brain template is centered (i.e., with minimal distance to all samples of its parent population) and discriminative (i.e., can easily distinguished from a different CBT generated from a different class). Let $\mathcal{D} = \{ \mathbf{X}_1, \dots, \mathbf{X}_n\}$ denote a population of brain connectivity matrices, where $\mathbf{X}_s \in \mathbb{R}^{r\times r}$ encodes the interaction weights between $r$ ROIs for a subject $s$. Here we leverage the state-of-the-art deep graph normalizer (DGN) network \cite{gurbuz2020deep} to learn an integral and holistic connectional template of the input population $\mathcal{D}$. DGN takes two or more weighted (possibly unweighted)  graphs and maps them onto an output population center graph (i.e., connectional template). This learning task is fundamentally rooted in mapping connectivity patterns onto a high-dimensional vector representation for each node in each graph in the given population, namely a node feature vector. During the mapping process, the unique domain-specific topological properties of the population graphs are preserved thanks to a topology-constrained normalization loss function which penalizes the deviation from the ground-truth population topology. Next, we derive the CBT edges from the pairwise relationship of node embeddings. The final CBT encoded in a matrix $\mathbf{C} \in \mathbb{R}^{r\times r}$ is learned by minimizing the Frobenius distance to a random set $S$ of individuals to avoid overfitting as follows:

\begin{equation*}
\argmin_{\mathbf{C}} \sum_{s \in S} \left \|  \mathbf{C} - \mathbf{X}_{s} \right \|_{F}  
\label{eq:1}
\end{equation*}

\textbf{Regression learning task (time-series brain graph evolution prediction).} Foreseeing the brain evolution \cite{gilmore2012longitudinal,mills2016structural,meng2014spatial} as a complex highly interconnected system is crucial for mapping dynamic interactions between different brain ROIs for different neurological states in the dataset. In fact, atypical alterations in the brain connectome evolution trajectory might indicate the emergence of neurological disorders such as Alzheimer's disease (AD) or dementia \cite{lohmeyer2020attitudes,Ezzine:2019}. Hence, learning to predict longitudinal brain dysconnectivity from a baseline timepoint helps to diagnose such disorders at an early stage. Such regression task aims at learning how to generate real brain graphs at follow-up timepoints from a single baseline graph acquired at timepoint $t_1$. Recently, EvoGraphNet \cite{nebli2020deep} was proposed as the first GNN for predicting the evolution of brain graphs at $t_i > t_1$ by cascading a set of time-dependent graph generative adversarial networks (gGANs), where the predicted graph at the current timepoint $t_i$ is passed on to train the next gGAN in the cascade, thereby generating the brain graph at timepoint $t_{i+1}$. To optimize the learning process of generation, EvographNet minimizes the adversarial loss between a pair of generator $G_i$ and descriminator $D_i$ at each time point $t_i$. EvoGraphNet architecture assumes sparse changes in the brain connectivity structure over time and includes $l1$ loss for that reason. Besides, it also introcuces $\mathcal{L}_{KL}$, a loss based on Kullback-Leibler (KL) divergence \cite{kldivergence}, to enforce the alignment between the predicted and ground-truth domains. The full loss is defined as follows:
\begin{equation}
    \mathcal{L}_{FUll} = \sum_{i=1}^{T}(\lambda_1 \mathcal{L}_{adv}(G_i, D_i) + \frac{\lambda_2}{n}\sum_{tr=1}^{n}\mathcal{L}_{l1}(G_i, t_r)
    + \frac{\lambda_3}{n}\sum_{tr=1}^{n}\mathcal{L}_{KL}(t_i, t_r)
    )      
\end{equation}
where $\lambda_1$, $\lambda_2$ and $\lambda_3$ are hyperparameters to moderate the importance of each component. For our one-representative shot learning, to train EvoGraphNet architecture with a single sample trajectory, we learn time-dependent CBTs from the available training samples at each timepoint $\{ t_i\}_{i=1}^T$, independently. Each gGAN in the cascade learns how to evolve the population CBT $\mathbf{C}^{t_i}$ from timepoint $t_i$ into the CBT $\mathbf{C}^{t_{i+1}}$ at the follow-up timepoint $t_{i+1}$.

\textbf{Diagnostic learning task (brain graph classification).} The graph classification task aims to predict the label of a graph instance. This task is performed heavily on chemical and social domains \cite{errica2019fair}. In brain graph classification, instances consist of brain connectomes. Graph attention networks (GAT) \cite{velivckovic2017graph} are one of the most powerful convolution-style neural networks that operate on graph-structured data. It uses masked self-attentional layers. Its performance does not depend on knowing the full graph structure and it can operate on different number of neighbours while assigning different importance to each node. Considering its performance and efficiency and our sample structure, we choose GAT to perform our experiments with brain graph classification task into different neurological states. In order to integrate one representative-shot learning approach with the brain graph classification task, we use one CBT per class. This means that every class is represented by one sample that is derived from its whole class population using DGN \cite{gurbuz2020deep}. Then GAT model is then trained on \emph{only two samples (i.e., class-specific templates)} in a supervised manner. Next, the trained method is fed with testing samples and according to the predicted possibility values, testing samples are labelled.  

Given that GAT is based on self-attention computations, we denote the shared attentional mechanism as $a : \mathbb{R}^{F'} \times \mathbb{R}^{F'} \rightarrow \mathbb{R}$ where $F$ and $F'$ are the dimensions of the node features in the current and the following layers, respectively. 
For each node $i$ in the graph, we compute the attention coefficients of its direct neighbors: 

\begin{equation*}
e_{ij} = a(\mathbf{W} \Vec{h_i}, \mathbf{W} \Vec{h_j}) 
\end{equation*}

where $\mathbf{W} \in{\mathbb{R}^{F \times F'}}$ is a learnt weight matrix and $\Vec{h_i} ,\Vec{h_j} \in{\mathbb{R}^{F}}$ denote the feature vectors of the nodes $i$ and $j$, respectively. The coefficients are then normalized using a softmax function: 

\begin{equation*}
\alpha_{ij} = softmax_j(e_{ij})=\frac{exp(e_{ij})}{\sum_{k \in \mathcal{N}_i} exp(e_{ik})}  
\end{equation*}

where $\mathcal{N}_i$ is the set of neighbors of node $i$ in the graph. Finally, the new node features are calculated using a non-linearity function: 
\begin{equation*}
\sigma: \Vec{h'_i}= \sigma (\sum_{j \in \mathcal{N}_i} \alpha_{ij} \mathbf{W} \Vec{h_j}) 
\end{equation*}

\section{Results and Discussion}

\subsection{Regression learning task}

\textbf{Longitudinal dataset.} We used $113$ subjects from the OASIS-2\footnote{\url{https://www.oasis-brains.org/}} longitudinal dataset \cite{Marcus:2010}. This set consists of a longitudinal collection of 150 subjects with a ranging age from  60 to 96. Each subject has 3 visits (i.e., timepoints), separated by at least one year. Each subject is represented by one cortical morphological brain network derived from T1-w MRI. Specifically, we use Desikan-Killiany cortical template to partition the cortical surface into 35 ROIs, each hemisphere (left and right). Next, by computing the absolute difference in the maximum principal curvature between pairs of ROIs we define the morphological weight between them.

\textbf{Parameters.} In order to find the CBT, we used 5-fold cross-validation strategy. We set the random training sample set $S$ size to 10, learning rate to 0.0005 and trained 100 epochs with early stopping criteria. For EvoGraphNet \cite{nebli2020deep}, we use adaptive learning rates for the generators decreasing from 0.01 to 0.1 while we fixed the discriminator learning rate to 0.0002. The $\lambda$ parameters for the loss functions are set to $\lambda_1 = 2$,  $\lambda_2 = 2$, and $\lambda_3 = 0.001$, respectively. We used the AdamW \cite{loshchilov2018fixing} optimizer with 0.5 and 0.999 moment estimates. We trained our model over 300 epochs using a single Tesla V100 GPU (NVIDIA GeForce GTX TITAN with 32GB memory). These hyperparameters were selected empirically using grid search. 

\textbf{Evaluation and comparison methods.} We compared our one-representative shot learning method against EvoGraphNet \cite{nebli2020deep} (trained on all samples) and two other one-shot learning strategies, namely (i) linear average one-shot and (ii) random one-shot. In the linear average one-shot learning, we created the CBT by simply linearly averaging the population samples at each timepoint, independently. For the random one-shot, we picked a single random sample to train on at baseline timepoint. We repeated the random sampling 20 times and report the average results. All methods used EvoGraphNet architecture with the set parameters. The first training strategy of EvoGraphNet presents an upper bound as it is trained with all samples. In \textbf{Table}~\ref{time_series_mae}, we report the mean absolute error (MAE) between the ground-truth and predicted graphs.

\begin{table}[ht!]
\begin{center}

\resizebox{\textwidth}{!}{\begin{tabular}{c | c | c | c | c}
\hline
& \multicolumn{2}{c|}{$t_1$} & \multicolumn{2}{c}{$t_2$} \\
\hline
\textbf{Method}   
&  \begin{tabular}{c c}
    Left Hemisphere
\end{tabular} 
& \begin{tabular}{c c}
    Right Hemisphere
\end{tabular}
&   \begin{tabular}{c c}
    Left Hemisphere
\end{tabular} 
&  \begin{tabular}{c c}
    Right Hemisphere
\end{tabular} \\
\hline
EvoGraphNet \cite{nebli2020deep} (all)  & $0.053 \pm 0.0069$ & $0.070 \pm 0.0221$ & $0.069 \pm 0.0149$ & $\mathbf{0.161 \pm 0.1208}$ \\
Random one-shot & $0.098 \pm 0.0051$ & $0.126 \pm 0.0268$ & $0.201 \pm 0.0168$ & $0.289 \pm 0.1454$ \\
Linear average one-shot & $0.078 \pm 0.0199$ & $0.064 \pm 0.0090$ & $0.183 \pm 0.0466$ & $0.212 \pm 0.1599$ \\
CBT one-shot (ours) & $\mathbf{0.047 \pm 0.0031}$ & $\mathbf{0.053 \pm 0.0035}$ & $\mathbf{0.056 \pm 0.0017}$ & $0.214 \pm 0.0759$ \\

\hline
\end{tabular}}
\caption{\label{time_series_mae} Mean absolute error (MAE) between ground-truth and predicted brain graphs at follow-up timepoints by our proposed method and comparison methods including standard deviations across different folds.}
\end{center}
\vspace{-6mm}
\end{table} 
 
Our method clearly outperformed both linear average and random one-shot learning methods. Remarkably, it also outperformed the EvoGraphNet except for the right hemispheric data at $t_2$. Altough EvoGraphNet (all) presents an upper bound in the expected performance, our method, while being trained on single sample, outperformed it in three out of four cases. The standard deviation is also higher in other methods especially in random one-shot learning. Comparing the CBT one-shot against the linear average one demonstrates that linear averaging does not produce a highly-representative template of the parent population, hence the low performance. This also implicitly shows the ability of DGN in learning well representative and centered CBTs that can be leveraged in downstream learning tasks. Our CBT one-shot method also achieved the highest stability against train and test distribution shifts via cross-validation by lowering the MAE standard deviation across folds. This clearly indicates that our single population-representative shot is enough to train a cascade of gGAN architectures without resorting to expensive augmentation and prolonged batch-based training. Overall, our method provided significant results in terms of accuracy and stability, and we even achieved a better performance than when training on all samples.

\subsection{Diagnostic learning task}

\textbf{Dataset.} We evaluated our CBT one-shot learning framework on a brain dementia dataset consisting of 77 subjects (41 subjects diagnosed with Alzheimer's diseases (AD) and 36 with Late Mild Cognitive Impairment (LMCI)) from the Alzheimer's Disease Neuroimaging Initiative (ADNI) database GO public dataset \cite{mueller2005alzheimer}. Each subject is represented by a morphological network derived from the maximum principle curvature of the cortical surface. 

\textbf{Parameters.} For CBT estimation, we used the same DGN parameter setting as in the previous section for the brain graph evolution prediction task. For the main architecture of GAT\cite{velivckovic2017graph}, we used a dropout rate of 0.6 and removed weak connections by thresholding each brain connectivity matrix using its mean value. We set the initial learning rate to 0.0001 and the weight decay to 0.0005. For Leaky ReLU, we set the alpha to 0.2. These hyperparameters were chosen empirically.

\textbf{Evaluation and comparison methods.}  We also used 5-fold cross-validation strategy for classification evaluation. We compared our method against GAT and both linear average and random one-shot learning strategies. All methods used GAT for brain graph classification. The classification results on both right and left brain graph datasets are detailed in \textbf{Table}~\ref{Tab:classification_results_right} and \textbf{Table}~\ref{Tab:classification_results_left}, respctively. In addition to regression, we evaluated our CBT one-shot training strategy using a classification GNN model (here GAT) to further demonstrate the \emph{generalizability} of our method \emph{across various learning tasks}. Using two different datasets, we ranked \emph{first} for the left hemisphere dataset and \emph{second best} for the right hemisphere in terms of classification accuracy. Both Tables \ref{Tab:classification_results_right} and \ref{Tab:classification_results_left} indicate that the specificity scores are low. However, our method brought way higher sensitivity which is more important for clinical applications as we need to avoid false negatives. We note that the brain changes are very subtle between LMCI and AD patients, hence the low classification accuracy across all methods given the difficulty of this task. In fact, the distribution of samples drawn from both classes are highly overlapping. Many studies have pointed out the difficulty of the discrimination task between LMCI and AD patients \cite{goryawala2015inclusion}. In fact, this task is considered as a hard problem in the AD literature which is least addressed relatively \cite{mahjoub2018brain}. We also notice that for the right hemisphere we had a highly competitive performance with the upper-bound method (GAT trained on all). Again, the best classification performance across both hemispheres was achieved using our CBT one-shot strategy. Our reproducible conclusive results across different learning tasks challenges the assumption of deep learning models being `data-hungry' and prone to failure when learning from a few samples. We have shown that it is possible to \emph{learn with one sample and achieve the best performance without any augmentation} if the sample is well-representative of the given population --recall that random and averaging cases did not perform well. Possible extensions of this novel line of research include the debunking of the learning behavior of GNN models in resourceful and frugal training conditions and eventually laying the theoretical principles of training such heavy networks on \emph{templates}.

\begin{table}[t]
\begin{center}
\begin{tabular}{|c|c|c|c|c|}
\hline
Method                                                 & Accuracy      & Sensitivity   & Specificity  & AUC           \\ \hline
GAT \cite{velivckovic2017graph} (all) & {\textbf{0.52}}    & 0.44          & {\ul{0.6}}   & {\textbf{0.52}}    \\ \hline
Linear average one-shot                                & 0.49          & 0.24          & \textbf{0.78} & \ul{0.51}          \\ \hline
Random one-shot                                        & 0.5           & {\ul{ 0.5}}    & 0.5         & 0.5          \\ \hline
CBT one-shot (ours)                                    & \ul{0.51} & \textbf{0.69} & 0.3          & 0.49 \\ \hline
\end{tabular}
\caption{AD/LMCI brain classification results by the proposed method and benchmarks for the right hemisphere dataset. 4 different metrics are used to compare: accuracy, sensitivity, specificity, AUC. Results are the average of the 5 folds. \textbf{Bold}: best. \uline{Underlined}: second best.}
\label{Tab:classification_results_right}
\end{center} 
\vspace{-10mm}
\end{table}

\begin{table}[h]
\begin{center}
\begin{tabular}{|c|c|c|c|c|}
\hline
Method                                                 & Accuracy      & Sensitivity   & Specificity  & AUC           \\ \hline
GAT \cite{velivckovic2017graph} (all) & {\ul{0.51}}    & 0.36          & {\ul{0.68}}   & {\ul{0.52}}    \\ \hline
Linear average one-shot                                & 0.42          & 0.18          & \textbf{0.7} & 0.44          \\ \hline
Random one-shot                                        & 0.5           & {\ul{ 0.49}}    & 0.51         & 0.5          \\ \hline
CBT one-shot (ours)                                    & \textbf{0.54} & \textbf{0.67} & 0.4          & \textbf{0.53} \\ \hline
\end{tabular}
\caption{AD/LMCI brain classification results by the proposed method and benchmarks for the left hemisphere dataset. 4 different metrics are used to compare: accuracy, sensitivity, specificity, AUC. Results are the average of the 5 folds. \textbf{Bold}: best. \uline{Underlined}: second best.}
\label{Tab:classification_results_left}
\end{center} 
\vspace{-10mm}
\end{table}

\section{Conclusion}

In this paper, we showed that different GNN architectures trained using a single representative training graph can outperform models trained on large datasets across different tasks. Specifically, our method uses population-driven templates (CBTs) for training. We have showed that our model generated promising results in two different tasks: brain graph evolution prediction and classification. The concept of training on a CBT has the potential to be generalized to other GNN models. Although many theoretical investigations have yet to be taken, leveraging a representative CBT for training such models may become a useful component in alleviating the costly batch-based training of GNNs models. The CBT one-shot learning opens a new chapter in enhancing the GNNs performance while potentially reducing their computational time by capturing the commonalities across brain connectivities populations via the template representation.

\section{Acknowledgements}

This work was funded by generous grants from the European H2020 Marie Sklodowska-Curie action (grant no. 101003403, \url{http://basira-lab.com/normnets/}) to I.R. and the Scientific and Technological Research Council of Turkey to I.R. under the TUBITAK 2232 Fellowship for Outstanding Researchers (no. 118C288, \url{http://basira-lab.com/reprime/}). However, all scientific contributions made in this project are owned and approved solely by the authors. M.A.G is supported by the same TUBITAK 2232 Fellowship.


\bibliography{Biblio}
\bibliographystyle{splncs}
\end{document}